\pdfoutput=1

\documentclass[11pt]{article}

\usepackage{ACL2023}

\usepackage{times}
\usepackage{latexsym}

\usepackage[T1]{fontenc}

\usepackage[utf8]{inputenc}

\usepackage{microtype}

\usepackage{inconsolata}

\usepackage{amsmath,amsfonts,bm}

\def\eqref#1{equation~\ref{#1}}

\def\1{\bm{1}}

\def\vx{{\bm{x}}}

\DeclareMathAlphabet{\mathsfit}{\encodingdefault}{\sfdefault}{m}{sl}
\SetMathAlphabet{\mathsfit}{bold}{\encodingdefault}{\sfdefault}{bx}{n}

\def\gC{{\mathcal{C}}}
\def\gD{{\mathcal{D}}}

\def\gX{{\mathcal{X}}}

\usepackage{url}
\usepackage{hyperref}
\usepackage{booktabs}
\usepackage{nicefrac}
\usepackage{xcolor}
\usepackage{graphicx}
\usepackage{cleveref}
\usepackage{wrapfig}
\usepackage{subcaption}
\usepackage{dirtytalk}
\usepackage{textcomp}
\usepackage{listings}
\usepackage{amsmath}
\usepackage{amssymb}
\usepackage{amsfonts}
\usepackage{makecell}
\usepackage{multicol}
\usepackage{multirow}
\usepackage{bbding}
\usepackage{colortbl}
\usepackage{pifont}
\usepackage{mathrsfs}
\usepackage{bbm}
\usepackage{bm}
\usepackage{enumitem}
\usepackage{xspace}
\RequirePackage{algorithm}
\RequirePackage{algorithmic}

\newcommand{\method}{\texttt{TTIDA}\xspace}
\title{\method: Controllable Generative Data Augmentation via \\ Text-to-Text and Text-to-Image Models}

\author{
  Yuwei Yin\textsuperscript{\rm 1}, 
  Jean Kaddour\textsuperscript{\rm 2}, 
  Xiang Zhang\textsuperscript{\rm 3}, 
  Yixin Nie\textsuperscript{\rm 4}, 
  \\ {\bf Zhenguang Liu}\textsuperscript{\rm 5}, 
  {\bf Lingpeng Kong}\textsuperscript{\rm 1}, 
  {\bf Qi Liu}\textsuperscript{\rm 1} 
  \\
  \textsuperscript{\rm 1} Department of Computer Science, University of Hong Kong; \textsuperscript{\rm 2} University College London\\
  \textsuperscript{\rm 3} University of Alberta; \textsuperscript{\rm 4} University of North Carolina at Chapel Hill; \textsuperscript{\rm 5} Zhejiang University\\
  \{ywyin, lpk, liuqi\}@cs.hku.hk; jean.kaddour.20@ucl.ac.uk; \\ xzhang23@ualberta.ca; yixin1@cs.unc.edu; zhenguangliu@zju.edu.cn
}

\begin{document}

\maketitle

\begin{abstract}
Data augmentation has been established as an efficacious approach to supplement useful information for low-resource datasets. Traditional augmentation techniques such as noise injection and image transformations have been widely used. In addition, generative data augmentation (GDA) has been shown to produce more diverse and flexible data. While generative adversarial networks (GANs) have been frequently used for GDA, they lack diversity and controllability compared to text-to-image diffusion models.
In this paper, we propose \method (\underline{T}ext-to-\underline{T}ext-to-\underline{I}mage \underline{D}ata \underline{A}ugmentation) to leverage the capabilities of large-scale pre-trained Text-to-Text (T2T) and Text-to-Image (T2I) generative models for data augmentation. By conditioning the T2I model on detailed descriptions produced by T2T models, we are able to generate photo-realistic labeled images in a flexible and controllable manner.
Experiments on in-domain classification, cross-domain classification, and image captioning tasks show consistent improvements over other data augmentation baselines. Analytical studies in varied settings, including few-shot, long-tail, and adversarial, further reinforce the effectiveness of \method in enhancing performance and increasing robustness.\footnote{\url{https://github.com/YuweiYin/TTIDA}}
\end{abstract}

\section{Introduction}
\label{sec:intro}

Data augmentation is ubiquitous in the preprocessing procedure of various machine learning tasks, particularly those involving insufficient labeled data \cite{image_da_survey}. It provides multiple benefits, such as minimizing the costs associated with collecting and annotating data, alleviating concerns related to data scarcity and imbalance, and mitigating the deleterious effects of overfitting on model generalization. The efficacy of data augmentation is predicated on the extent to which the augmented dataset approximates the underlying data distribution. Thus the optimal objective is to obtain a wide range of samples that reflect the natural distribution of richness and diversity in a given target object category.

Most conventional augmentation methods employed in computer vision rely on handcrafted transformations that utilize a restricted array of elementary invariances, such as rotation, cropping, and color adjustment \cite{image_da_survey, da_dl_img_clf, adaptive_img_da}. These transformations are pre-specified and applied uniformly across the entire dataset, which may not be optimal for different types of data or scenarios. Such a limitation motivates the need for more advanced and flexible approaches to augment data effectively in the context of various computer vision tasks.
To obtain more diverse images, prior studies \cite{da_gan, gan_aug} on generative data augmentation (GDA) aim to approximate the data distribution of the observed image dataset using generative adversarial networks \cite{gan} (GANs).
However, each category needs a separate GAN model to be trained, which is inflexible and incurs considerable training expenses. Besides, the training process of GANs is notoriously unstable, especially when the training set is small. It also brings the problem of mode collapse, which produces images with less diversity.

In this paper, we aim to address the limitations of existing GDA methods for vision tasks by exploring publicly accessible text-to-image (T2I) models, which generate images according to the input text. These models are based on diffusion models \cite{diffusion_model, stable_diffusion} and have been trained with large-scale text-image pairs obtained from the Web.
Hence T2I models can generate a variety of photo-realistic images that are conditioned on diverse text descriptions. The benefits of utilizing these models are manifold. 1) they provide a language interface that enables flexible control and generation of desired images; 2) they are domain-agnostic and capable of generating extensive data with high diversity; 3) they serve as a versatile tool to synthesize high-quality data for vision tasks in various scenarios, including in-domain, cross-domain, long-tail, and so forth.

In the field of natural language processing, \citet{gpt_text_da} have utilized the large pretrained text-to-text (T2T) model GPT-2 \cite{gpt2} as a method of data augmentation for text classification tasks.
Building on this concept, we propose \method (Text-to-Text-to-Image Data Augmentation) to leverage the generative capabilities of large-scale pretrained text-to-image (T2I) and text-to-text (T2T) models for the purpose of data augmentation. Specifically, we first fine-tune a T2T model, such as GPT-2 and T5 \cite{t5}, on a diverse set of image captions. Then, for each object category, the label text is inputted to the T2T model to obtain a more detailed description of the object of our desire.
Following this, a T2I model, such as GLIDE \cite{glide}, is employed to generate multiple photo-realistic images of the object, which are conditioned on either the original label text or the more detailed description. The synthetic images thus produced are utilized to augment the original image dataset.
Using detailed descriptions as prompts for the T2I model is beneficial as label text typically contains only one or two words and is, therefore, typically simplistic.
Therefore, the use of T2T models brings controllability to the generated images, and the diversity of generated text also guarantees the diversity of generated images.

To evaluate the efficacy of our approach, a series of experiments are conducted on various tasks, including (1) image classification with distinct scenarios such as balanced, long-tail, and adversarial data settings, (2) cross-domain image classification, and (3) image captioning. 
The experimental results on CIFAR \cite{cifar}, Office \cite{office_31, office_home}, and MS COCO \cite{mscoco} benchmarks substantiate the consistent performance enhancement of our method compared to other data augmentation baselines. Remarkably, we observe that the superiority of our approach is more notable in instances of scarce data or diverse data domains. Furthermore, we show that \method is capable of enhancing model robustness in the face of adversarial attacks.
Our contributions are summarized as follows.

\begin{itemize}[itemsep=0pt]
    \item We propose \method, a novel approach combining the generation power of large-scale pre-trained text-to-text (T2T) and text-to-image (T2I) models for data augmentation in a controllable and flexible way.
    \item We demonstrate the efficacy of \method in enhancing the model performance on in-domain classification, cross-domain classification, and image captioning benchmarks.
    \item The analytical studies in varied settings, including few-shot, long-tail, and adversarial, further reinforce the effectiveness and robustness of \method.
\end{itemize}

\section{Related Work}
\label{sec:related_work}

\subsection{Traditional Image Transformations}
The application of conventional augmentation techniques for computer vision has been widely acknowledged and validated to be effective \cite{image_da_effectiveness}. These techniques typically rely on manually crafted transformations that exploit a restricted set of elementary invariances \cite{goodfellow2016deep}, including but not limited to cropping, rotation, and flipping \cite{image_da_survey}.
The mere inclusion of a greater number of image augmentation techniques does not invariably culminate in an improvement in performance since certain methods may demonstrate a sensitivity to the selection of image augmentations \cite{byol}.

\subsection{Generative Data Augmentation}
\citet{da_gan} propose to use image-conditional Generative Adversarial Networks (GANs) \cite{gan} to generate within-class images conditioned on a source domain. Similarly, \citet{gan_aug} use GANs to augment data for brain segmentation tasks.
In addition to GANs, some researchers utilize the pretrained language model GPT-2 \cite{gpt2} as a method of data augmentation for multiple text classification tasks \citet{gpt_text_da} and commonsense reasoning benchmarks \cite{gda_commonsense}.
Besides generative approaches, another line of work tries to learn advanced augmentation strategies using a fixed subset of augmentation functions. \citet{auto_augment, rand_augment} use reinforcement learning to automatically find a dataset-specific augmentation policy which is proved to be effective on the downstream tasks.

\begin{figure*}[ht]
    \centering
    \includegraphics[width=0.7\textwidth]{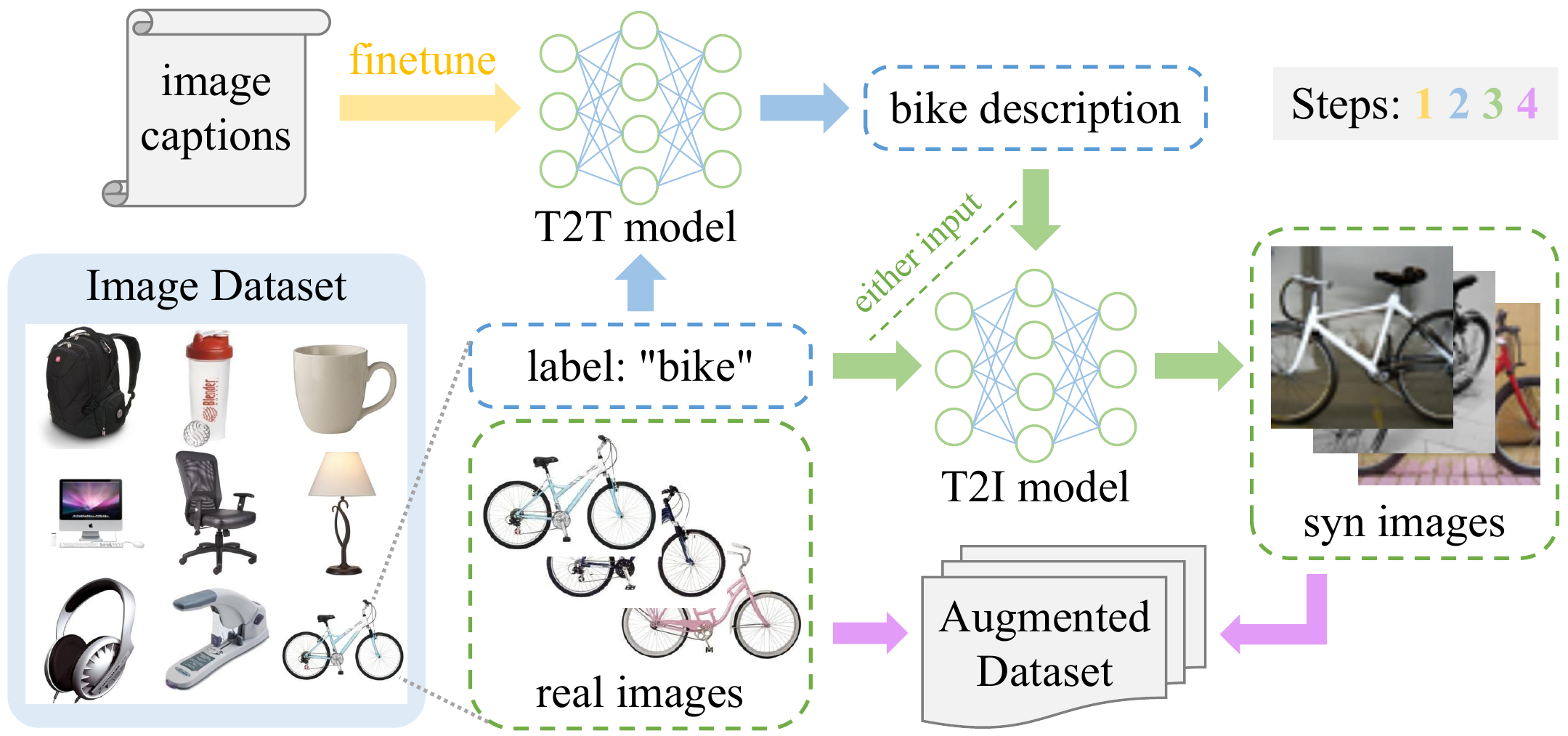}
    \caption{\textbf{Overview of \method} (Text-to-Text-to-Image Data Augmentation). Arrows in different colors represent different steps. \textcolor{orange}{Step 1}: finetune the text-to-text (T2T) model using text captions of images. \textcolor{cyan}{Step 2}: input the label text, i.e., ``bike'', to the T2T model to produce a caption-like description of bikes. \textcolor{green}{Step 3}: input either the original label text or the generated description into the text-to-image (T2I) model to generate high-quality synthetic (syn) images. \textcolor{purple}{Step 4}: combine the real images from the original dataset with the augmented images for model training.}
    \label{fig:overview}
\end{figure*}

\section{Method}
\label{sec:method}

In this section, we elaborate on our \method approach to generative data augmentation and the motivations the behind our framework.

\subsection{Overview}
\label{subsec:method_overview}

Figure \ref{fig:overview} shows the overview of our data augmentation method, where arrows in different colors denote different pipeline steps. For each object category, i.e., bike in Figure \ref{fig:overview}, we input the label text ``bike'' to the T2I model such as GLIDE \cite{glide} to generate multiple photo-realistic images of this object (\textcolor{green}{Step 3}). Then we combine the real images from the original dataset with the generated synthetic images together (\textcolor{purple}{Step 4}). The augmented dataset is directly used for model training.
Usually, the label text is a word or short phrase. To automatically obtain a finer prompt for the T2I model, we can first input the label text to a text-to-text (T2T) generative model finetuned with image captions (\textcolor{orange}{Step 1}) to produce a longer object description (\textcolor{cyan}{Step 2}), e.g., ``a white bike near the wall''.
Step 1 and Step 2 are optional since the T2I model can still generate high-quality images with the label text input. Yet the T2T model can produce precise or personalized object descriptions with a richer context, increasing the diversity of synthetic images to a large extent.

\subsection{Formulation}
\label{subsec:method_formulation}

We denote the training set of an image classification dataset as $\gD = \{ \gD_1, \dots, \gD_n \}$ that has $n$ different categories. The $i$-th category is $\gD_i = \{ l_i, \gX_i \}$, where $l_i$ is the label text, and $\gX_i = \{ \vx_1, \dots, \vx_m \}$ are $m$ real images in this category.
\textcolor{orange}{Firstly}, we finetune the T2T model $\texttt{t2t}(\cdot)$ on the caption corpus with the language modeling loss (next-token prediction) conditioned on the corresponding prompt $\mathrm{p}$, i.e., to maximize:
\begin{equation}\label{equ:language_modeling}
    \sum_u \sum_v \log p(t_v^u | t_{<v}^u; \mathrm{p}^u),
\end{equation} where $u$ is the number of captions in the training set of image captioning datasets, and $v$ is the token index of the $u$-th caption. $t_{<v}^u$ denotes all the previous generated tokens before the $v$-th token. The prompt $\mathrm{p}^u$ of the $u$-th caption includes the caption's entities.

\textcolor{cyan}{Then}, we input the category label $l_i$ to the T2T model to obtain the description sentence:
\begin{equation}
    d_i = \texttt{t2t}(l_i).
\end{equation}

\textcolor{green}{After that}, the T2I model $\texttt{t2i}(\cdot)$ uses either $l_i$ or $d_i$ as the input prompt to generate plenty of diverse synthetic images $\{ \hat{\gX}_i = \hat{\vx}_i^j \}_{j=1}^{G}$ of the same object as depicted in $l_i$ or $d_i$, where $G$ is the number of synthetic images:
\begin{align}
    & \text{either} & \hat{\vx}_i^j = \texttt{t2i}(l_i) \\
    & \text{or} & \hat{\vx}_i^j = \texttt{t2i}(d_i).
\end{align}

\textcolor{purple}{Lastly}, we merge the original data of the $i$-th category $\gD_i = \{ l_i, \gX_i \}$ with the generated images $\hat{\gX}_i$ to construct the augmented category $\gD_i^{aug}$:
\begin{equation}
    \gD_i^{aug} := \{ l_i, \gX_i, \hat{\gX}_i \} = \gD_i \cup \hat{\gX}_i.
\end{equation}

Repeat this process for every category $\gD_i$ in the dataset $\gD$, we can have a augmented dataset $\gD^{aug} = \{ \gD_1^{aug}, \dots, \gD_n^{aug} \}$.

\subsection{Text-to-Text Models}
\label{subsec:method_t2t}

We use Generative Pre-Training (GPT-2) \cite{gpt2} as the T2T model. Specifically, we adopt the basic \texttt{GPT2LMHeadModel}\footnote{\url{https://huggingface.co/gpt2}} and set the maximal sentence length as $20$ and the beam size as $5$ for beam search. The GPT model is finetuned with the language modeling loss, and Adam optimizer \cite{adam} for $5$ epochs using all the captions in the training set of MS COCO 2015 Image Captioning Task \cite{mscoco}.

When fine-tuning GPT-2, the prompt $\mathrm{p}^u$ in Equation \ref{equ:language_modeling} is a template containing all entity tokens $\{e_i^u\}_{i=1}^{n}$ extracted from the $u$-th caption sentence beforehand. We prepend the template ``\textit{Write an image description with keywords including $e_1^u$, $e_2^u$, $\dots$, and $e_n^u$:}'' to the original caption sentence. Then we feed the prompted caption into the GPT-2 model to finetune GPT-2 using the language modeling loss.
In this way, by editing the entity words of input prompts when generating image captions using the fine-tuned GPT-2 model, we can flexibly control the generated sentences and thus can modify the contents of the generated images by the T2I model.

\subsection{Text-to-Image Models}
\label{subsec:method_t2i}

We adopt Guided Language to Image Diffusion for Generation and Editing (GLIDE) \cite{glide} as our T2I model.
Unlike other text-to-image models that mainly focus on the generation of pictures with different artistic styles, such as DALL-E \cite{dalle2} and Stable Diffusion \cite{stable_diffusion}, GLIDE aims to generate photo-realistic pictures from the input prompt text. We feed the T2I model with image description sentences produced by the T2T model to generate synthetic images as data augmentation in a more controllable manner.

\section{Experimental Setup}
\label{sec:exp_conf}
In this section, we elaborate on all the experimental setups, including task introduction (\ref{subsec:task_intro}), datasets (\ref{subsec:datasets}), data augmentation (\ref{subsec:data_augmentation}), our method and baseline methods (\ref{subsec:our_baseline}), backbone models (\ref{subsec:backbone_models}). See Appendix \ref{app:training_details} for detailed training details.

\subsection{Tasks}
\label{subsec:task_intro}
To illustrate the versatility of our method, we tackle a diverse set of computer vision tasks. We now explain on a case-by-case basis how these settings can demonstrate the benefits of \method.

\paragraph{In-domain Image classification} In this task, the domain of the training set is the same as that of the test set. We consider a balanced data settings: Each category has the same number of images. Other settings such as Few-shot, long-tail, and adversarial will be discussed in Section \ref{sec:analysis}.

\paragraph{Cross-domain Image classification} The model trains on the images of one domain and tests on those of another. For example, a cross-domain dataset $\gD$ has $K$ domains $\gD = \{ \gD_1, \dots, \gD_K \}$, and each domain $\gD_k$ contains the same $n$ categories $\{ \gC_1^k, \dots, \gC_n^k \}$. The images in the $i$-th category $\gC_{src}^i$ of the source domain $\gD_{src}$ and those in the $i$-th category $\gC_{tgt}^i$ of the target domain $\gD_{tgt}$ denote the same object. Images of different domains are visibly different in some aspects, such as angle and position, tone of color, and background features. 
Hence, experimental results on this task will demonstrate the benefits of generating diverse images and thus verify the domain-agnostic trait of our method.

\paragraph{Image Captioning} In this task, models need to generate a caption sentence that describes the input image accurately. By testing the performance of \method on this task, we show the advantages of combining the power of autoregressive language models (T2T models) and T2I models to generate reasonable (text, image) pairs for improving the image-to-text generation (captioning) ability.

\subsection{Datasets}
\label{subsec:datasets}

\paragraph{In-domain Image classification} We conduct in-domain image classification experiments on two datasets. CIFAR-100 contains $100$ different distinguishable classes and a smaller size of training samples--$500$ images--in each class. We use the classification accuracy as the evaluation metric.

\paragraph{Cross-domain Image classification} We adopt two cross-domain datasets to measure the effectiveness of \method. 1) The Office-31 dataset \cite{office_31} contains $31$ object categories in three domains: \textit{Amazon}, \textit{DSLR} and \textit{Webcam}.
2) Office-Home \cite{office_home} is another benchmark dataset for domain adaptation containing $4$ domains including \textit{Art}, \textit{Clipart}, \textit{Product}, and \textit{Real-World}, where each domain has $65$ categories.

\paragraph{Image Captioning} We use the image captioning dataset of the Microsoft COCO (common objects in context) 2015\footnote{\url{https://cocodataset.org/\#captions-2015}} Image Captioning Task. MS COCO Captions \cite{mscoco} contains over one and a half million captions describing over $330,000$ images. For the training and validation images, five independent human-generated captions are provided for each image.

\subsection{Data Augmentation}
\label{subsec:data_augmentation}

\paragraph{Synthetic Images for CIFAR} The images in the CIFAR dataset have the size of $32 \times 32$; simply specifying the generation with such low resolution from GLIDE will largely decrease the quality of images. Instead, we generate images of size $256 \times 256$ with more details and then perform a resizing. For CIFAR-100, $500$ images are produced by the GLIDE model. In total, we provide $50000$ additional training samples for these two datasets.

\paragraph{Synthetic Images for Office} Unlike CIFAR, the original images in the Office-31 and Office-Home datasets vary from low resolution to high resolution. Besides, these images are not in the same shape, so CDTrans performs a series of image transformations. Our synthetic images are of size $256 \times 256$ initially and then perform the same transformations as in CDTrans. For each category in each domain, the number of synthetic images generated by T2I models is the same as that of Office datasets.

\paragraph{Synthetic Images for MS COCO} We generate synthetic images using the training set of COCO captions as augmentation. In addition, we extract entities from COCO captions using CoreNLP\footnote{\url{https://stanfordnlp.github.io/CoreNLP/}} and Natural Language Toolkit (NLTK)\footnote{\url{https://www.nltk.org/}}, and then feed the T2I model with the entities or synthetic captions generated by T2T models using these entities.

\subsection{Our Method and Baseline Methods}
\label{subsec:our_baseline}

\paragraph{\method}
We use the T2T and T2I models as described in Section \ref{subsec:method_t2t} and \ref{subsec:method_t2i}. Specifically, for all experiments, we use the basic model settings of the public GLIDE model\footnote{\url{https://github.com/openai/glide-text2im}} and follow the standard generation procedure: we use the \texttt{base} GLIDE model to generate $64 \times 64$ images, and then feed them into the \texttt{upsample} model to obtain high-quality images of $256 \times 256$ resolution. Then we perform a resizing to match the image size of different datasets. Except for resizing and image normalization, no other image transformations are adopted.

\paragraph{Traditional Image Transformations}
The incorporation of additional image augmentation techniques does not inevitably lead to improved performance, as certain methods may be sensitive to the choice of augmentations \cite{byol}. Thus we adopt the image transformation procedure\footnote{\url{https://github.com/facebookresearch/moco-v3}} employed in SimCLR \cite{simclr} and MoCo \cite{moco_v1, moco_v2, moco_v3}, which has been validated for its efficacy.

\paragraph{GAN-based Generative Data Augmentation}
To contrast our \method approach with previous generative data augmentation methods, we employ three representative GAN models, namely DCGAN \cite{dcgan}, CycleGAN \cite{cyclegan}, and StyleGAN \cite{stylegan_v1, stylegan_v2, stylegan_v3}, to augment images. More specifically, we adhere to the original implementations of DCGAN\footnote{\url{https://github.com/pytorch/examples/tree/main/dcgan}}, CycleGAN\footnote{\url{https://github.com/junyanz/CycleGAN}}, and StyleGAN\footnote{\url{https://github.com/NVlabs/stylegan3}} for GAN training and generation.

\subsection{Backbone Models}
\label{subsec:backbone_models}

\paragraph{In-domain Image Classification} For all experiments on CIFAR, we use the standard ResNet-101 architecture \cite{resnet} as the backbone\footnote{\url{https://pytorch.org/vision/main/models/generated/torchvision.models.resnet101.html}}.

\paragraph{Cross-domain Image Classification} We adopt the state-of-the-art model CDTrans \cite{cdtrans} as the baseline model. We follow the implementation of CDTrans\footnote{\url{https://github.com/CDTrans/CDTrans}} and only add a data processing module for incorporating our synthetic images into the original source-domain dataset.

\paragraph{Image Captioning} We use the state-of-the-art model mPLUG \cite{mplug} as our baseline and follow the implementation of mPLUG\footnote{\url{https://github.com/alibaba/AliceMind/tree/main/mPLUG}}. Similarly, we add a data processing block for integrating our augmented data into the COCO captioning dataset.

\section{Results}
\label{sec:exp_res}

In this section, we report all experimental results w.r.t. the settings described in Section \ref{sec:exp_conf}.

\begin{table}[t]
\small
\centering
\scalebox{0.9}{
\begin{tabular}{c|cccc}
\toprule
CIFAR-100 & + 20\% & + 50\% & + 100\% & + Max \\
\midrule
Img Trans DA & +0.3\% & +0.4\% & +0.4\% & +0.5\% \\
DCGAN DA & +0.4\% & +0.5\% & +0.7\% & +1.0\% \\
CycleGAN DA & +0.5\% & +0.7\% & +1.0\% & +1.2\% \\
StyleGAN DA & +0.7\% & +0.9\% & +1.2\% & +1.4\% \\
\method (label) & +1.1\% & +1.8\% & +2.3\% & +2.7\% \\
\method (desc.) & \textbf{+1.3\%} & \textbf{+2.1\%} & \textbf{+2.6\%} & \textbf{+3.0\%} \\
\bottomrule
\end{tabular}
}
\caption{\textbf{Classification accuracy on CIFAR-100}. We report accuracy improvements when adding synthetic images generated by different models described in Section \ref{subsec:our_baseline}. ``+Max'' denotes the highest score using $200\%$, $300\%$, $400\%$, or $500\%$ synthetic images.
}
\label{tab:exp_cifar_100}
\end{table}

\subsection{In-domain Image Classification}

The experimental results of image classification on the CIFAR-100 dataset are shown in Table \ref{tab:exp_cifar_100}. \method outperforms all baselines on each synthetic ratio, demonstrating the effectiveness of our method in boosting the image classification performance.

\subsection{Cross-domain Image Classification}

\begin{table}[t]
\small
\centering
\scalebox{0.95}{
\begin{tabular}{c|ccc}
\toprule
Office-31 & A$\to$D & A$\to$W & D$\to$A \\
\midrule
Before Tuning & 97.0\% & 96.7\% & 81.1\% \\
w/o Syn Data & 97.4\% & 96.8\% & 81.3\% \\
\textbf{w/ Syn Data} & \textbf{98.0\%} & \textbf{97.1\%} & \textbf{81.6\%} \\
\midrule
Office-31 & D$\to$W & W$\to$A & W$\to$D \\
\midrule
Before Tuning & 99.0\% & 81.9\% & 100\% \\
w/o Syn Data & 99.0\% & 82.0\% & 100\% \\
\textbf{w/ Syn Data} & \textbf{99.1\%} & \textbf{82.2\%} & \textbf{100\%} \\
\bottomrule
\end{tabular}
}
\caption{\textbf{Target-domain classification accuracy in every direction on the Office-31 dataset.} Office-31 has three domains, namely Amazon (A), DSLR (D), and Webcam (W). ``Before Tuning'' stands for the test scores of the best checkpoints of the state-of-the-art model CDTrans. ``w/ Syn Data'' and ``w/o Syn Data'' represent the results after finetuning with and without synthetic images generated by \method respectively.
}
\label{tab:exp_office_31}
\end{table}

\begin{table}[t]
\small
\centering
\scalebox{0.95}{
\begin{tabular}{c|ccc|c}
\toprule
Art & A$\to$C & A$\to$P & A$\to$R & Avg A \\
\midrule
Before Tuning & 68.8\% & 85.0\% & 86.9\% & 80.23\% \\
w/o Syn Data & 68.9\% & 85.4\% & 87.1\% & 80.47\% \\
\textbf{w/ Syn Data} & \textbf{69.2\%} & \textbf{85.7\%} & \textbf{87.6\%} & \textbf{80.83\%} \\
\midrule
Clipart & C$\to$A & C$\to$P & C$\to$R & Avg C \\
\midrule
Before Tuning & 81.5\% & 87.1\% & 87.3\% & 85.30\% \\
w/o Syn Data & 81.8\% & 87.2\% & 87.4\% & 85.47\% \\
\textbf{w/ Syn Data} & \textbf{82.2\%} & \textbf{87.5\%} & \textbf{87.4\%} & \textbf{85.70\%} \\
\midrule
Product & P$\to$A & P$\to$C & P$\to$R & Avg P \\
\midrule
Before Tuning & 79.6\% & 63.3\% & 88.2\% & 77.03\% \\
w/o Syn Data & 79.7\% & 64.5\% & 88.3\% & 77.50\% \\
\textbf{w/ Syn Data} & \textbf{80.1\%} & \textbf{65.9\%} & \textbf{88.5\%} & \textbf{78.17\%} \\
\midrule
RealWorld & R$\to$A & R$\to$C & R$\to$P & Avg R \\
\midrule
Before Tuning & 82.0\% & 66.0\% & 90.6\% & 79.53\% \\
w/o Syn Data & 82.6\% & 66.1\% & 90.7\% & 79.80\% \\
\textbf{w/ Syn Data} & \textbf{82.8\%} & \textbf{66.4\%} & \textbf{90.9\%} & \textbf{80.03\%} \\
\bottomrule
\end{tabular}
}
\caption{\textbf{Target-domain classification accuracy in every direction on the Office-Home dataset.} Office-Home has four domains, namely Art (A), Clipart (C), Product (P), and RealWorld (R). ``Before Tuning'' stands for the test scores of the best CDTrans checkpoints. ``w/ Syn Data'' and ``w/o Syn Data'' represent the results after finetuning for $50$ epochs with and without synthetic images generated by \method, respectively.
}
\label{tab:exp_office_home}
\end{table}

We continue training the best CDTrans checkpoints for $50$ epochs on every Office domain adaption direction with and without our synthetic images. The target-domain classification results on Office-31 and Office-Home datasets are reported in Table \ref{tab:exp_office_31} and Table \ref{tab:exp_office_home}, respectively. We observe that fine-tuning consistently enhances classification accuracy for all directions, especially when training with our augmented data. The results verify the effectiveness of \method in improving model performance on cross-domain image classification tasks.

\subsection{Image Captioning}

\begin{table}[ht]
\small
\centering
\scalebox{0.98}{
\begin{tabular}{cc|ccc}
\toprule
\multicolumn{2}{c}{\#Data + Syn Ratio} & BLEU4 & ROUGE-L & CIDEr-D \\
\midrule
5000 & + 100\% & +3.2\% & +1.0\% & +2.4\% \\
10000 & + 100\% & +1.9\% & +2.1\% & +3.0\% \\
50000 & + 100\% & +1.5\% & +1.3\% & +5.2\% \\
100000 & + 100\% & +0.5\% & +0.2\% & +1.0\% \\
200000 & + 100\% & +0.7\% & +0.3\% & +2.1\% \\
\bottomrule
\end{tabular}
}
\caption{\textbf{Test scores of mPLUG model finetuned on MS COCO 2015 Image Captioning dataset}. We report BLEU, ROUGE, and CIDEr scores with (+ 100\%) and without (+ 0) synthetic images generated by \method.
}
\label{tab:exp_mscoco_img_cap}
\end{table}

We train the base mPLUG \cite{mplug} model (\texttt{mplug.en.base}) based on pre-trained CLIP \cite{clip} model (\texttt{ViT-B-16}) for $5$ epochs on the COCO image captioning dataset of different training size. The model is evaluated using BLEU \cite{bleu}, ROUGE \cite{rouge}, and CIDEr \cite{cider} metrics.
Specifically, we adopt the evaluator implementation \cite{mscoco_eval} to calculate BLEU4, sentence-level ROUGE-L, and CIDEr-D scores\footnote{\url{https://github.com/tylin/coco-caption}}.

Table \ref{tab:exp_mscoco_img_cap} compares the test scores of the finetuned mPLUG model with and without the synthetic data generated by \method. As the results show, \method can further boost the model performance on different evaluations under different data size settings.

\begin{figure}[ht]
    \centering
    \includegraphics[width=0.95\linewidth]{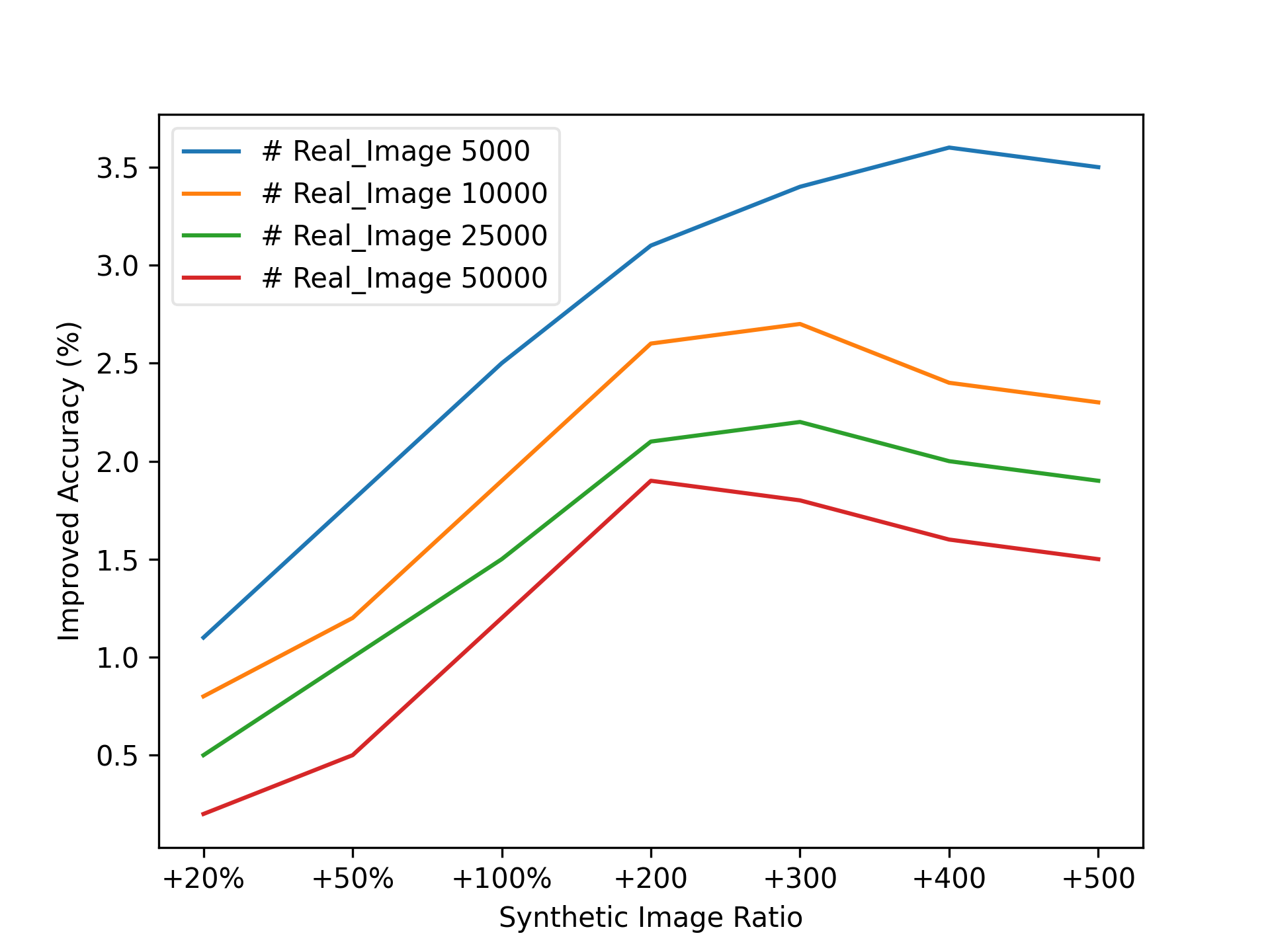}
    \caption{Results on the CIFAR-100 benchmark under the few-shot setting.}
    \label{fig:cifar100_few_shot}
\end{figure}

\section{Analysis}
\label{sec:analysis}
In this section, we conduct analytic studies to better understand how our proposed framework contributes to the model performance.

\subsection{Synthetic Images of Different Ratios}

Commonly, data augmentation is more useful in low-resource settings than in high-resource ones. To verify the idea, we proportionately adjust the scale of the training set to create high- and low-resource contexts. Augmentation of the original training set with synthetic images of varying ratios is executed with the aim of determining the optimal conditions for the application of \method. The experimental results of image classification on the CIFAR-100 dataset are shown in Figure \ref{fig:cifar100_few_shot}. We report the classification accuracy (\%) on the CIFAR datasets with synthetic training images of different proportions to the total number of original images.

It reveals that the incorporation of a greater number of synthetic images leads to a discernible reduction in classification error across all cases. Additionally, it is observed that the efficacy of the proposed method is particularly prominent in situations where the quantity of original CIFAR training images is meager. These findings suggest that the augmentation strategy proposed in this work holds particular promise in cases where the original dataset is characterized by limited resources. To further bolster this claim, the efficacy of \method is evaluated on synthetic long-tail CIFAR datasets.

\subsection{Training on Long-tail Datasets}

\begin{figure}[ht]
    \centering
    \includegraphics[width=0.95\linewidth]{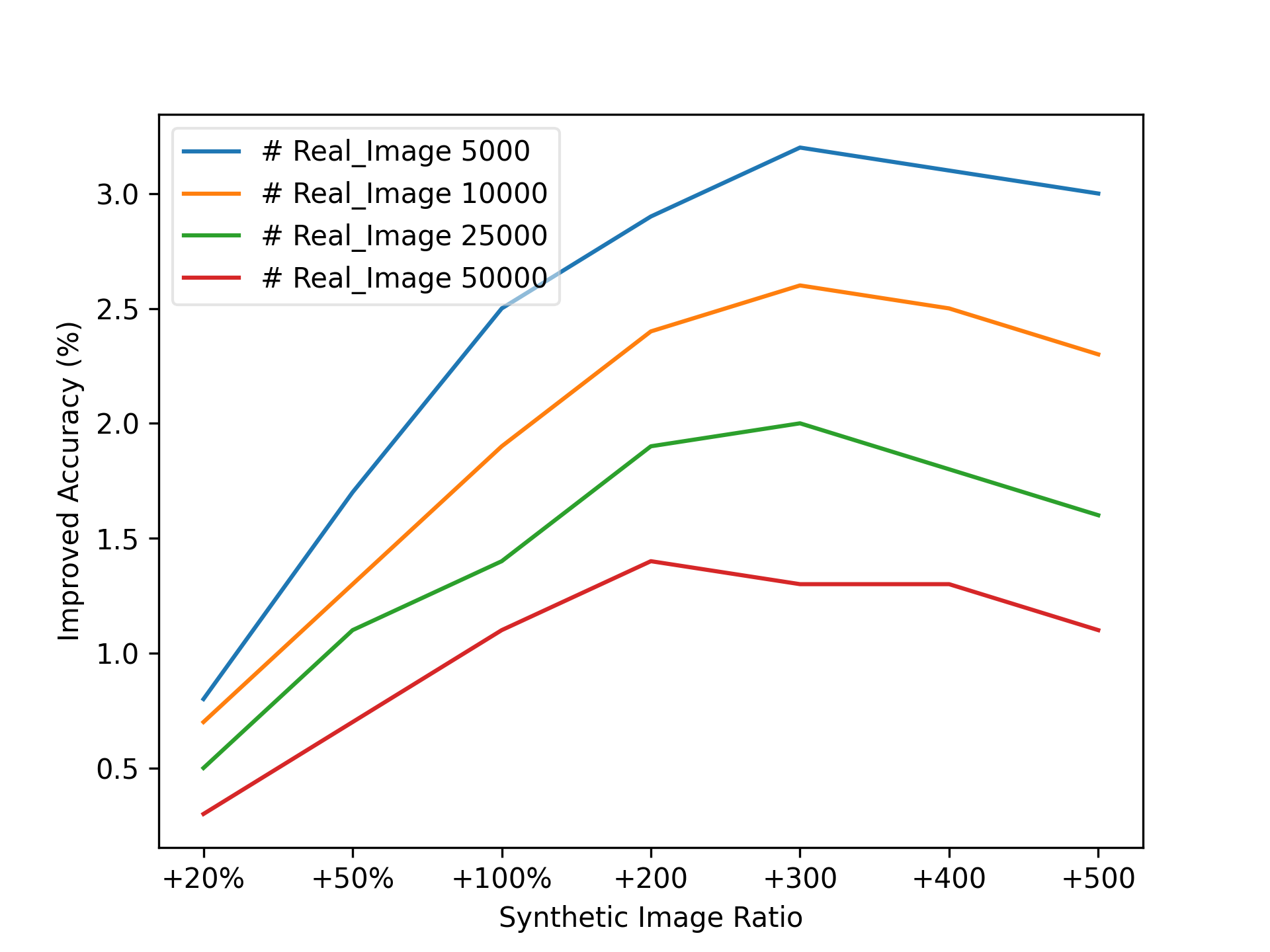}
    \caption{Results on the Long-tail CIFAR-100 benchmark under the few-shot setting.}
    \label{fig:cifar100_few_shot_lt}
\end{figure}

We test our method on the synthetic long-tail CIFAR subset, where each category has a different number of images. We construct the Long-tail CIFAR subset $\gD^{lt} = \{ \gD_1^{lt}, \dots, \gD_n^{lt} \}$ from the original balanced CIFAR dataset $\gD = \{ \gD_1, \dots, \gD_n \}$ that has as $n$ different categories, where the $i$-th sub-category $\gD_i^{lt}$ has $i/n$ of the images in $\gD_i$.

We report the improvement of classification accuracy (\%) after using synthetic images generated by \method on the long-tail CIFAR-100 dataset. As shown in Figure \ref{fig:cifar100_few_shot_lt}, the experimental results are consistent with the findings in Figure \ref{fig:cifar100_few_shot}. Thus we conclude that the effectiveness of \method in improving the accuracy of image classification, especially in low-resource cases, is further verified.

\subsection{Training with Adversarial Images}

\begin{table}[t]
\small
\centering
\scalebox{0.95}{
\begin{tabular}{c|cc|c}
\toprule
Dataset & + 100\% Syn & + Adv & Acc \\
\midrule
CIFAR-100 &  & \ding{52} & $-$1.8\% \\
CIFAR-100 & \ding{52} & \ding{52} & $-$0.7\% \\
\bottomrule
\end{tabular}
}
\caption{\textbf{Classification accuracy on CIFAR-100 with adversarial training images.} We set the number of original training images to $10$k and report the results with and w/o synthetic (Syn) and adversarial (Adv) data added to the training set.
}
\label{tab:exp_adversarial}
\end{table}

We conduct experiments after adding adversarial images to the training set. For each class, we manually collect such images from the Internet, e.g., images with unusual styles (Appendix \ref{app:case_study}). This experiment will test whether a diverse set of synthetic data helps to make the classifier more robust.

As shown in Table \ref{tab:exp_adversarial}, we compare the differences in classification accuracy before and after adding the adversarial images to the test set. The model trained with augmented data performs $2.80$\% better, demonstrating that it boosts model robustness to unusual images.

\begin{table}[ht]
\small
\centering
\scalebox{0.95}{
\begin{tabular}{c|cc|c}
\toprule
Dataset & label & description & Acc \\
\midrule
CIFAR-100 & \ding{52} &  & +1.7\% \\
CIFAR-100 &  & \ding{52} & +1.9\% \\
\midrule
Dataset & label & description & Avg Acc \\
\midrule
Office-31 & \ding{52} &  & +0.25\% \\
Office-31 &  & \ding{52} & +0.35\% \\
\midrule
Dataset & label & description & BLEU4 \\
\midrule
MS COCO & \ding{52} &  & +1.5\% \\
MS COCO &  & \ding{52} & +2.1\% \\
\bottomrule
\end{tabular}
}
\caption{\textbf{Comparison of T2I prompts}. We compare the test scores (CIFAR-100, Office-31, MS COCO) after augmenting with generated images with the original label text or descriptions generated by \method.}
\label{tab:exp_ablation_syn_type}
\end{table}

\subsection{Prompts Generated by T2T Models}

As shown in Figure \ref{fig:overview}, we can prompt the T2I model by either just a short label name or longer descriptions about the to-be-generated image. In this ablation, we compare the model performance between using the original label text or caption-like descriptions as prompts for the T2I model.

Table \ref{tab:exp_ablation_syn_type} shows that using rich descriptions to generate images results in a better improvement than using label text, especially on MS COCO benchmark. We hypothesize that using descriptions is more beneficial to cross-domain tasks and the large-scale COCO task since sentences produced by T2T models have richer context so that the T2I model can generate more diverse images. By contract, augmented data with too much diversity is less effective for the in-domain classification task CIFAR.

\subsection{Finetuning T2I Models}

To further explain the observations and verify the hypothesis in the former section, we finetune the T2I model on the training set of CIFAR-100 so that the synthetic images look more similar to the original images. Specifically, we follow the open-sourced practice\footnote{\url{https://github.com/afiaka87/glide-finetune}} to finetune GLIDE using the training set of MS COCO image captioning dataset. Table \ref{tab:exp_ablation_finetune_t2i} demonstrates that finetuning the T2I model brings further gains as expected.

\begin{table}[ht]
\small
\centering
\scalebox{0.95}{
\begin{tabular}{c|ccc|c}
\toprule
Dataset & label & description & finetune & Acc \\
\midrule
CIFAR-100 & \ding{52} &  &  & +1.5\% \\
CIFAR-100 & \ding{52} &  & \ding{52} & +2.0\% \\
\midrule
CIFAR-100 &  & \ding{52} &  & +1.9\% \\
CIFAR-100 &  & \ding{52} & \ding{52} & +2.3\% \\
\bottomrule
\end{tabular}
}
\caption{
\textbf{Comparison of finetuning T2I models}.
We compare the classification accuracy between using the original label text or descriptions on CIFAR-100 image classification task. The text-to-image model is finetuned on the CIFAR-100 training set if ``finetune'' is checked.
}
\label{tab:exp_ablation_finetune_t2i}
\end{table}

\section{Future Work}
\label{sec:future_work}

\subsection{Multimodal Back-Translation}

In machine translation, back-translation \cite{back_translation} is widely used as a powerful data augmentation method.
It performs a source$\to$target$\to$source translation process to augment the text data of the source language.
Similarly, a multimodal back-translation can be conducted using a text-to-image model like GLIDE and an image-to-text model like the model used in image captioning tasks.
Given available high-quality generators, we can generate extra text and image data by performing text$\to$image$\to$text and image$\to$text$\to$image back-translation respectively.

\subsection{Spurious Correlations}
Spurious correlations are correlations between image features and certain target classes where the features do not cause the latter \cite{causal_ml_survey}. They naturally occur in many datasets \cite{neuhaus2022spurious, vasudevan2022does, lynch2023spawrious}, but a model relying on them becomes problematic when faced with images where these correlations do not hold.
Given the flexibility of \method, we believe a promising direction is to augment training sets with images containing non-spurious features. This could be an attractive alternative for mitigating a model's reliance on spurious features compared to the current paradigm of downsampling majority groups  \cite{idrissi2022simple, schwartz2022limitations, throwing_away_data}.

\section{Conclusion}
\label{sec:conclusion}

In this work, we propose \method, a data augmentation scheme using large-scale pre-trained text-to-image and text-to-text models. We focus on generating photo-realistic images based on (i) concise label text and (ii) longer description prompts. Experimental results conducted for image classification and image captioning tasks under different settings demonstrate the effectiveness and robustness of the proposed augmentation method. Furthermore, we ablate some of our method's components and conclude with future work directions.

\section{Limitations}
\label{sec:limitations}

Despite the effectiveness of our method on the tasks above, this data augmentation approach can be applied to more machine learning applications, especially in low-resource situations. More experiments on different tasks with a wide range of hyper-parameter searching can further solidify our conclusions.
Besides, it is imperative to explore filtering strategies to avoid inappropriate data being added to the training set, e.g., outlier samples that are too dissimilar to the training set distribution according to some distance measure.

\section{Ethics Statement}
\label{sec:ethics_statement}

As described in \cite[Section 6]{glide}, the original GLIDE model generates fake but realistic images with possible disinformation or biases introduced by the data it was trained on. However, we believe that such ethical concerns can be addressed by appropriate data filtering, as suggested by \citet{glide}.


\bibliography{acl2023}
\bibliographystyle{acl_natbib}

\clearpage
\newpage

\appendix

\section{Case Study}
\label{app:case_study}

\begin{figure}[ht]
    \centering
    \begin{subfigure}[ht]{0.7\linewidth}
        \centering
        \includegraphics[width=\textwidth]{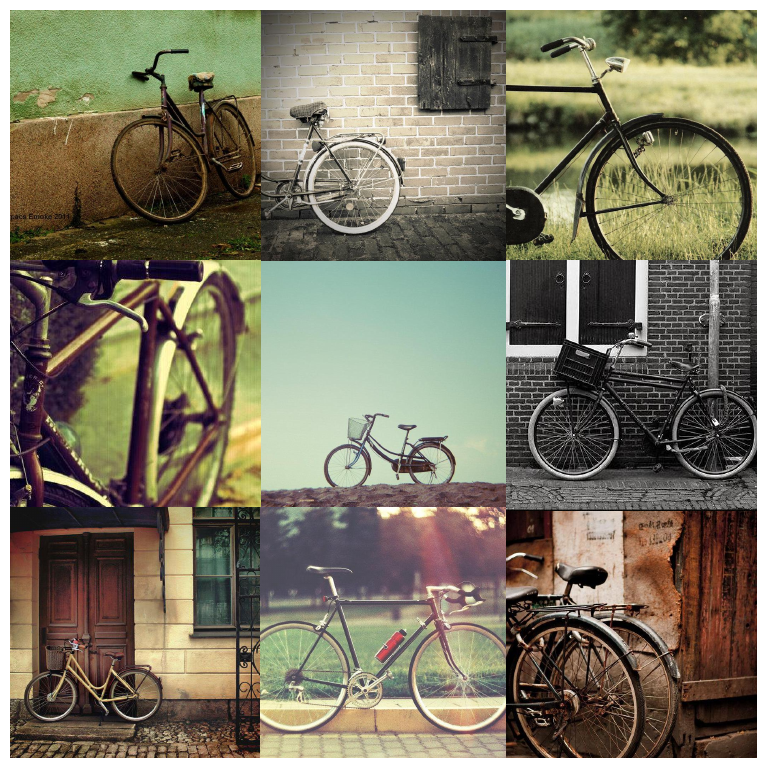}
        \caption{Bikes in Office-Home \textit{Art}.}
        \label{fig:office_home_art}
    \end{subfigure}
    \begin{subfigure}[ht]{0.7\linewidth}
        \centering
        \includegraphics[width=\textwidth]{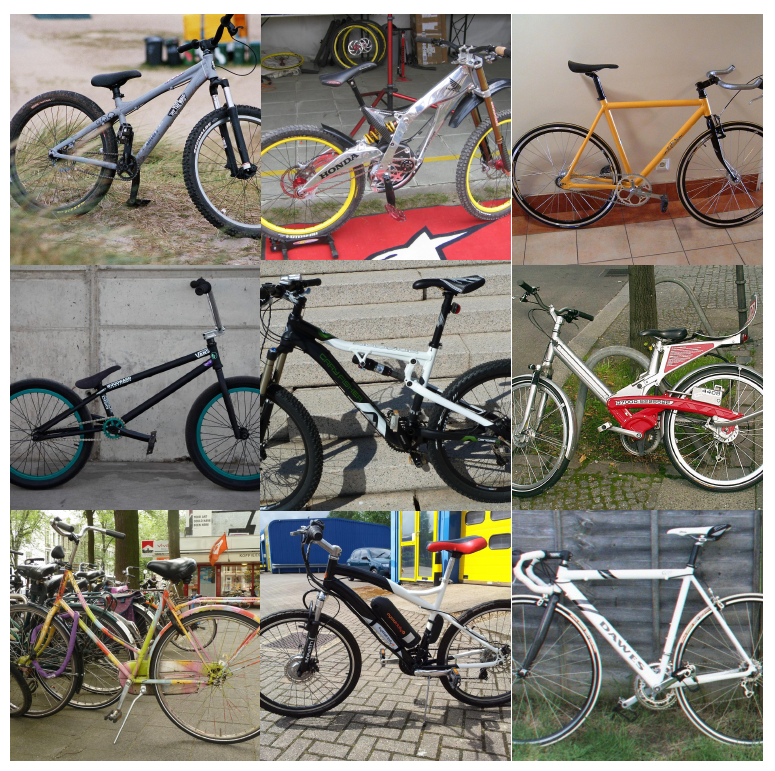}
        \caption{Bikes in Office-Home \textit{Real-World}.}
        \label{fig:office_home_real_world}
    \end{subfigure}
    \caption{Original bike images in the \textit{Art} and \textit{Real-World} domain of Office-Home dataset.}
    \label{fig:office_images}
\end{figure}

\paragraph{Original Images} Figure \ref{fig:office_images} shows the original bike images of 
the \textit{Art} and \textit{Real-World} domain of Office-Home dataset.
The image backgrounds and styles are visibly different in different domains.

\begin{figure}[ht]
    \centering
    \includegraphics[width=0.7\linewidth]{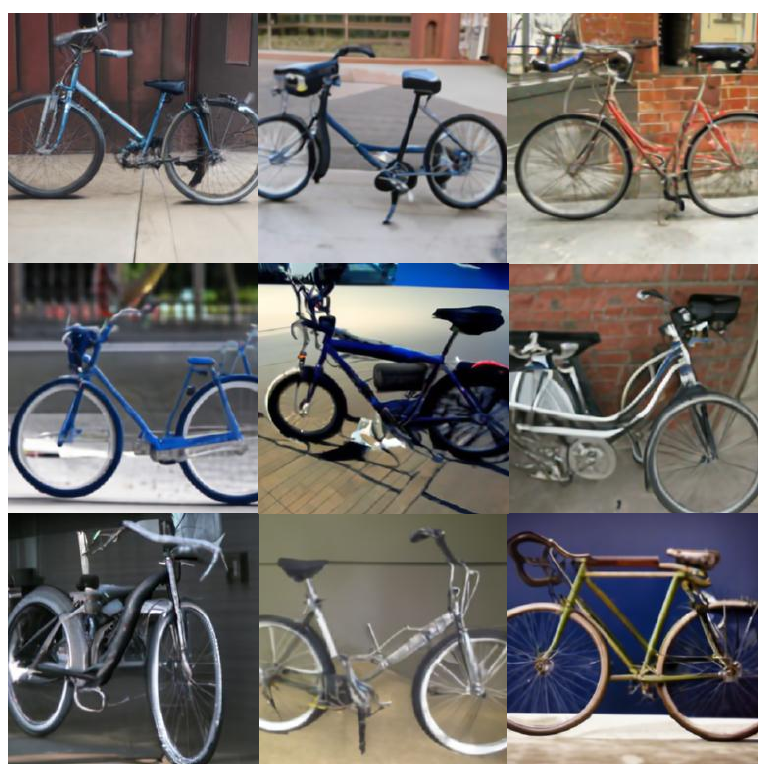}
    \caption{Synthetic bike images generated by \method.}
    \label{fig:synthetic_images}
\end{figure}

\paragraph{Our Synthetic Images} Figure \ref{fig:synthetic_images} shows the synthetic bike images generated by \method. We can observe that the vanilla GLIDE model is prone to generate images resembling photo-realistic images in the real world.

\begin{figure}[ht]
    \centering
    \includegraphics[width=0.7\linewidth]{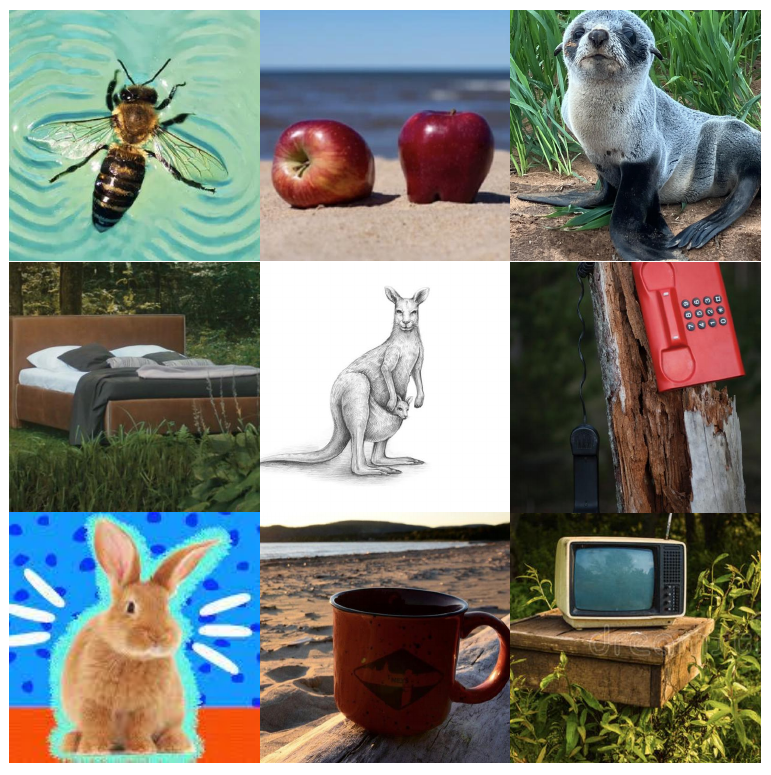}
    \caption{Adversarial images collected from the Internet.}
    \label{fig:adversarial_images}
\end{figure}

\paragraph{Adversarial Images} Figure \ref{fig:adversarial_images} shows the adversarial images we collected from the Internet. These images have unusual backgrounds or strange styles, so they break the spurious correlations between the content and background in the original dataset.

\section{Data Statistics}
\label{app:data_statistics}

Table \ref{tab:data_statistics_img_clf} lists the statistics of the datasets in in-domain (CIFAR) and cross-domain (Office) image classification tasks. 

\begin{table}[ht]
\small
\centering
\scalebox{0.9}{
\begin{tabular}{c|ccc}
\toprule
Dataset (\textit{Domain}) & \# total & \# class & \# per class \\
\midrule
CIFAR-100 & $50000$ & $100$ & $500$ \\
\midrule
Office-31 (\textit{Amazon}) & $2817$ & $31$ & $91$ \\
Office-31 (\textit{DSLR}) & $498$ & $31$ & $16$ \\
Office-31 (\textit{Webcam}) & $795$ & $31$ & $26$ \\
\midrule
Office-Home (\textit{Art}) & $2427$ & $65$ & $37$ \\
Office-Home (\textit{Clipart}) & $4365$ & $65$ & $67$ \\
Office-Home (\textit{Product}) & $4439$ & $65$ & $68$ \\
Office-Home (\textit{Real-World}) & $4357$ & $65$ & $67$ \\
\bottomrule
\end{tabular}
}
\caption{Statistics of image classification datasets. CIFAR-100 has 50000 images for training and extra 10000 images for testing. Office-31 dataset has 3 different domains and Office-Home dataset has 4 domains.}
\label{tab:data_statistics_img_clf}
\end{table}

\section{Training Details}
\label{app:training_details}

\paragraph{In-domain Image Classification} The standard ResNet-101 model is trained from inception for a duration of $200$ epochs on a single NVIDIA A40 GPU. To evaluate the model's efficacy, a holdout validation set is used, which is randomly sampled from every category of the training set by $20$ percent and is assessed after each epoch. Post-training, the optimal checkpoint is determined by selecting the best-performing model on the validation set. The selected checkpoint is then utilized to evaluate the accuracy of the model on the test set. For each setting, we repeat the process trice with three different random seeds $\{ 7, 17, 42 \}$ and report the average test score. 

For all classification experiments on CIFAR, the loss function is cross-entropy, and the batch size is $128$. We use the stochastic gradient descent (SGD) optimizer with an initial learning rate of $0.1$, momentum of $0.9$, and weight decay of $0.0005$. We perform a multi-step learning rate scheduler with gamma as $0.2$ and milestones as $\{ 60, 120, 160 \}$. In addition, the first $10$ epochs are warm-up epochs with linearly increasing learning rates.

\paragraph{Cross-domain Image Classification} We follow all the training settings of CDTrans \cite{cdtrans} and finetune the published best checkpoint on the Office benchmark for $50$ epochs.

\paragraph{Image Captioning} Similarly, We follow all the training details of mPLUG \cite{mplug} and finetune the published best checkpoint on the MS COCO training set for $5$ epochs.

\end{document}